\def\year{2020}\relax
\documentclass[letterpaper]{article} 
\usepackage{aaai20}  
\usepackage{times}  
\usepackage{helvet} 
\usepackage{courier}  
\usepackage[hyphens]{url}  
\usepackage{graphicx} 
\usepackage{graphicx}  
\usepackage{amsfonts}
\usepackage{booktabs}
\usepackage{multirow}
\usepackage{amsmath,bm}

\title{TreeGen: A Tree-Based Transformer Architecture for Code Generation}
\author{
        Zeyu Sun$^{\dag}$\ \ 
        Qihao Zhu$^{\dag}$\ \ 
        Yingfei Xiong\thanks{Yingfei Xiong is the corresponding author. The code is available at {\tt https://github.com/zysszy/TreeGen}}$^{\dag}$\ \ 
        Yican Sun$^{\dag}$\ \ 
        Lili Mou$^{\ddag}$\ \ 
        Lu Zhang$^{\dag}$\\
        $^{\dag}$Key Laboratory of High Confidence Software Technologies
(Peking University), MoE;\\
Software Institute, Peking University, 100871, P. R. China\\
        \{szy\_, zhuqh, xiongyf, sycpku, zhanglucs\}@pku.edu.cn\\
          $^{\ddag}$University of Alberta, Edmonton, AB, Canada\\
        doublepower.mou@gmail.com
    }

\begin{document}
\maketitle
\begin{abstract}
A code generation system generates programming language code based on an input natural language description. State-of-the-art approaches rely on neural networks for code generation. However, these code generators suffer from two problems. One is the long dependency problem, where a code element often depends on another far-away code element. A variable reference, for example, depends on its definition, which may appear quite a few lines before. The other problem is structure modeling, as programs contain rich structural information. 
In this paper, we propose a novel tree-based neural architecture, TreeGen, for code generation. TreeGen uses the attention mechanism of Transformers to alleviate the long-dependency problem, and introduces a novel AST reader (encoder) to incorporate grammar rules and  AST structures into the network. We evaluated TreeGen on a Python benchmark, HearthStone, and two semantic parsing benchmarks, ATIS and GEO. TreeGen outperformed the previous state-of-the-art approach by 4.5 percentage points on HearthStone, and achieved the best accuracy among neural network-based approaches on ATIS (89.1\%) and GEO (89.6\%). We also conducted an ablation test to better understand each component of our model.
\end{abstract}

\section{Introduction}
    Code generation is an important artificial intelligence problem that has the potential to significantly boost the productivity of programmers. Given a specification written in natural language, a code generation system translates the specification into an executable program.  
    For example, if a python programmer gives an instruction ``initialize a dictionary, Dict'', the code generator is expected to automatically generates ``\texttt{Dict=}\{\ \}''.
    
    With the development deep learning techniques, researchers have applied various neural architectures to this problem, such as sequence-to-sequence (Seq2Seq) models or sequence-to-tree (Seq2Tree) models~\cite{sutskever2014sequence,ling2016latent,yin2017syntactic,rabinovich2017abstract,hayati-etal-2018-retrieval,sun2019grammar}. 
    Especially, state-of-the-art approaches generate code by predicting a sequence of grammar rules~\cite{yin2017syntactic,rabinovich2017abstract,sun2019grammar}.
    That is to say, the system keeps a partial abstract syntax tree (AST) of the already-generated code, and predicts the grammar rule to be used to expand a particular node.
    
    The classification of grammar rules faces two main challenges. The first challenge is the long-dependency problem~\cite{bengio1994learning}. A code element may depend on another far-away element. For example, a variable reference statement  ``$\mathrm{if}\ \mathrm{len(a)} < \mathrm{Max\_Length}\text{:}$'' at line 100 may depend on a variable definition statement ``$\mathrm{Max\_Length}=100$'' at line 10. 
    The second challenge is the representation of code structures. It is pointed out that the tree-structural information is crucial for modeling code~\cite{mou2016convolutional,yin2017syntactic,rabinovich2017abstract,sun2019grammar}. 
     However, a ``flat'' neural architecture, such as an RNN, cannot capture structure information well.

    In this paper, we propose a novel neural architecture, TreeGen, for the code generation. To address the first challenge, TreeGen adopts the recently proposed Transformer architecture~\cite{vaswani2017attention}, which is capable of capturing long dependencies. However, the original Transformer architecture is not designed for programs, and cannot utilize tree structures, i.e., the second above mentioned challenge. A standard way of utilizing structural information, as in graph- and tree-based convolutional neural networks, is to combine the vector representations of a node and its structural neighbors as the output of a structural convolution sub-layer. However, a standard Transformer architecture does not have such structural convolution sub-layers, and it is not clear where to add them. 
    
    It is tempting to add structural convolution sub-layers in all the Transformer blocks.
    Our core conjecture is that when convolving a node and its structural neighbors, the vector representation should mainly contain the information from the original node. As the vector representation of the nodes is processed by more blocks in the decoder of the Transformer, they gradually mix in more information from other nodes and lose their original information. Therefore, we add the structural convolution sub-layer only to the first several Transformer decoder blocks but not all.

    Generally speaking, the TreeGen architecture 
    consists of three parts: (1) a natural language (NL) reader (encoder) encodes the text description; (2) an AST reader (the first several Transformer decoder blocks) encodes the previously generated partial code with the structural convolution sub-layers; (3) a decoder (the rest Transformer decoder blocks) combines the query (the node to be expanded in AST) and the previous two encoders to predict the next grammar rule. 

    We evaluated our model on an established benchmark dataset for Python code generation, HearthStone~\cite{ling2016latent}, which is a Python implementation of a card game HearthStone. The results show that our model significantly outperforms previous models by $4.5$ percentage points. We further evaluated our model on two semantic parsing datasets, ATIS and GEO, which translate natural language sentences into lambda calculus logical forms. The results show that our model has the best accuracy among previous neural models, with $89.1\%$ and $89.6\%$ accuracy, respectively.
    Our evaluation also shows that adding the structural convolution sub-layer to the first several Transformer blocks significantly outperforms a Transformer with structural convolution in all blocks.

\section{Our Model}
    \begin{figure}
        \centering
          \includegraphics[width=\linewidth]{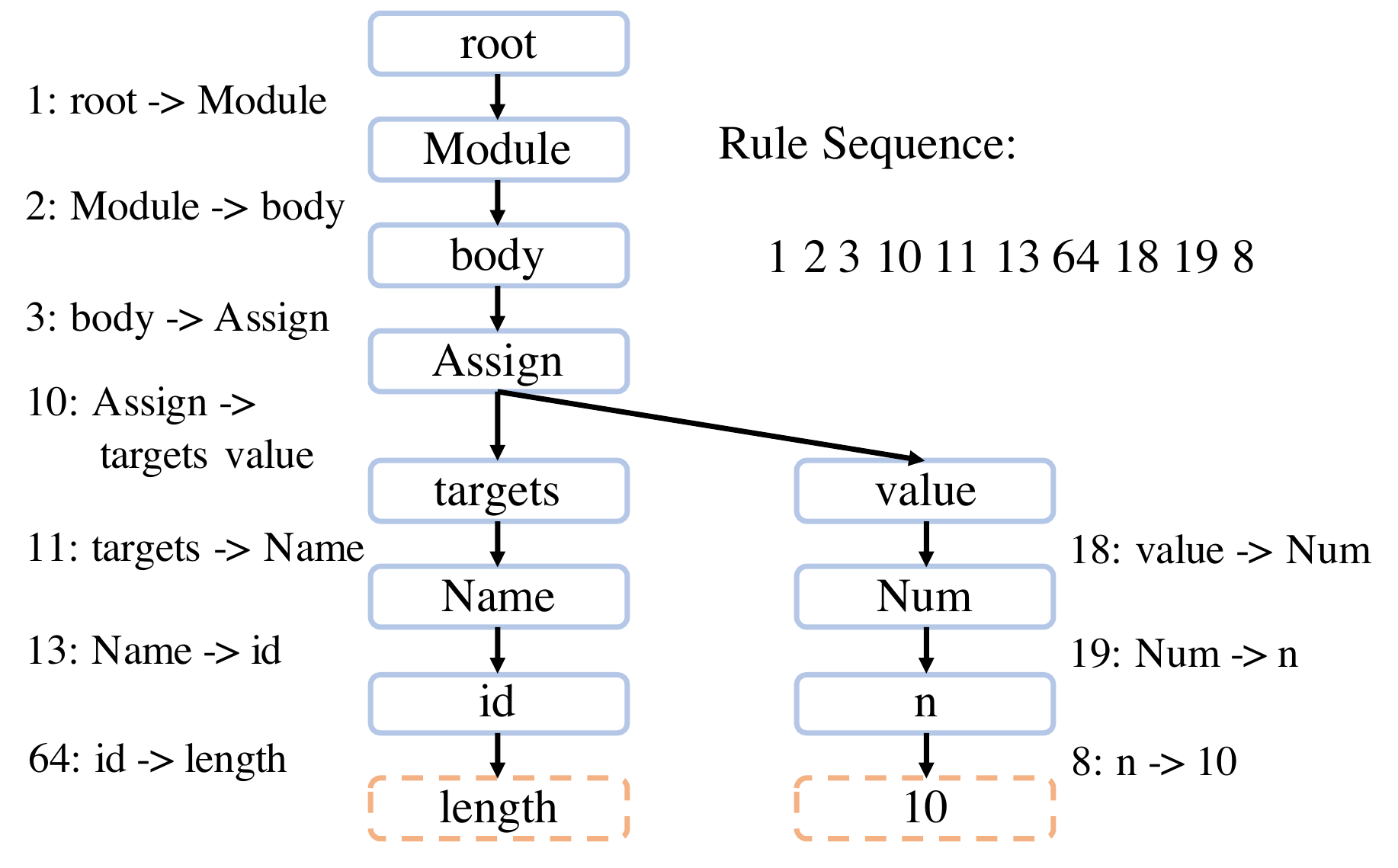}
        \caption{A Python AST for code ``$\mathrm{length} = 10$''}
        \label{AST_Example}
    \end{figure}
    \begin{figure*}
        \centering
          \includegraphics[width=\textwidth]{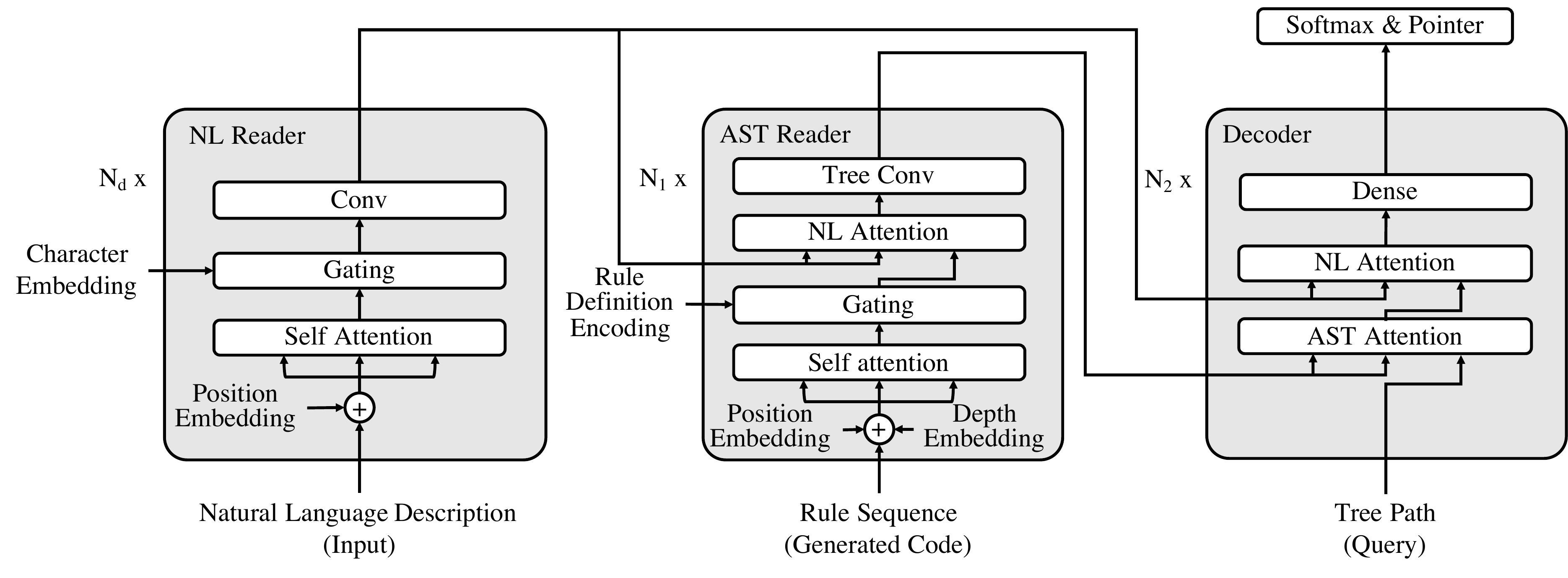}
        \caption{Overview of the TreeGen.}
        \label{Overview}
    \end{figure*}
    
    We generate code by predicting the grammar rules of the programming language. 
    Figure~\ref{Overview} shows the overall picture of our model, which comprises three parts: an NL reader, an AST reader, and decoder. We introduce them in detail in the following subsections.
    \subsection{Grammar Rule Prediction}

In this section, we introduce how to model code generation as a series of classification problems of grammar rules. The programs can be decomposed into several context-free grammar rules and parsed as an AST. For example, Figure~\ref{AST_Example} shows a Python AST for the code ``$\mathrm{length} = 10$'', where dotted boxes are terminal symbols and solid boxes are non-terminal symbols. 

AST-based code generation could be thought of as expanding a non-terminal node by a grammar rule. This process is repeated until all leaf nodes are terminal. 
In Figure~\ref{AST_Example}, ``\textit{1: root -$>$ Module}'' is an example of the grammar rules, where the preceding number is the ID of the rules. Following the pre-order traverse, we could obtain the sequence of rules that generate the AST shown in the top right corner. 

Formally, the probability can be factorized as the probabilities of the rules generating the code following the order.
    \begin{equation}
        p(\mathrm{code})=\prod\nolimits_{i=1}^P p(r_i\mid \text{NL input}, r_i, \cdots, r_{i-1})
    \end{equation}
where $r_i$ is the $i$th rule in the rule sequence. In this way, our task is to train a model to calculate $p(r_i \mid  \text{NL input}, p_i)$, i.e., given the natural language description and the currently generated partial AST the model calculates the probabilities of the rules to expand this node.

    \subsection{NL Reader}
        The input description determines the functionality of the code. It can be a semi-structural description as in the HearthStone dataset, or a natural language as in ATIS and GEO semantic parsing datasets.
        
        For an input description, we first tokenize it into tokens $n_1, n_2, \cdots, n_L$, where $L$ denotes the length of the input. Each token $n_i$ is then split to characters $c^{(n_i)}_1, c^{(n_i)}_2, \cdots, c^{(n_i)}_S$, where $S$ is the number of characters in $n_i$. All the tokens and characters are represented as real-valued vectors $\bm{n}_1, \bm{n}_2, \cdots, \bm{n}_L$ and $\bm{c}^{(n_i)}_1, \bm{c}^{(n_i)}_2, \cdots, \bm{c}^{(n_i)}_S$ by \textit{embeddings}.
        \subsubsection{Input Text Representation}\ 
        
            \medskip
            \underline{\textit{Character Embedding.}} It often happens that similar words have similar characters (e.g., ``program'' and ``programs''). 
            To utilize this property, we represent a token by character embeddings with a fully-connected layer
            \begin{equation}
                \bm{n}_i^\text{(c)} = W^\text{(c)}[\bm{c}^{(n_i)}_1;\cdots;\bm{c}^{(n_i)}_{M}]
                \label{equation:conv_preprocessing}
            \end{equation}
            where $W^\text{(c)}$ are the weights and the character sequence is padded to a pre-defined maximum length $M$.
            After the fully-connected layer, we also apply layer normalization~\cite{lei2016layer}. These vectors are then fed to the NL reader, and are integrated with the word embeddings by a gating sub-layer.
        
        \subsubsection{Neural Structure of NL Reader.} 
        The NL reader is composed of a stack of blocks ($N_d$ blocks in total). Each block contains three different sub-layers (namely, self-attention, gating mechanism, and word convolution) to extract features, which we introduce in detail in the following subsections. Between two sub-layers, we employ a residual connection~\cite{he2016deep} followed by a layer normalization. 
        
            \medskip
            \underline{\textit{Self-Attention.}} The self-attention sub-layer follows the Transformer's architecture~\cite{vaswani2017attention}, and uses multi-head attention to capture long dependency information. 

            For a sequence of input tokens $n_1, n_2, \cdots, n_L$, we represent them as an embedding 
            $\bm{n}_1, \bm{n}_2, \cdots, \bm{n}_L$ by a look-up table. We also use position embeddings to encode the information of word positions. In particular, we adopt the variant in~\citeauthor{dehghani2018universal}~(\citeyear{dehghani2018universal}), and compute the position embedding for the $i$th word in the $b$th Transformer block as 
            \begin{align}
            p_{b,i}[2j] &= \sin ((i+b)/(10000^{2j/d} ))\\
            p_{b,i}[2j + 1] &= \sin ((i+b)/(10000^{2j/d} ))
              \label{equation:position_embedding}
            \end{align}
            where $p_{i,b}[\cdot]$ indexes a dimension of the vector $\bm p_{i,b}$, and $d$ is the number of dimensions (i.e., embedding size). 

            
 A Transformer block learns non-linear features by multi-head attention, which yields a matrix $Y_b^\text{(self)} = [\bm{y}^\text{(self)}_{b, 1}, \bm{y}^\text{(self)}_{b, 2}, \cdots, \bm{y}^\text{(self)}_{b, L}]^\top$, where $Y_b^\text{(self)} \in \mathbb{R}^{L\times d}$. For notational simplicity, we omit the subscript $b$. The multi-head layer is computed by 
            \begin{equation}
                 Y^\text{(self)} =  \mathrm{concat}(head_1,\cdots,head_H) W_h
                \label{equation:head}
            \end{equation}
            where $H$ denotes the number of heads and $W_h$ is the weight. An attention layer is applied in each head $head_t$, computed by
            \begin{equation}
                head_t = \operatorname{softmax}(\frac{QK^\top}{\sqrt{d_k}})V
                \label{equation:dot_attention}
            \end{equation}
            where $d_k = d / H$ denotes the length of each features vector. $Q$, $K$ and $V$ are computed by 
            \begin{align}
            [Q, K, V]=[\bm x_1, \cdots, \bm x_L]^\top [ W_Q, W_K, W_V]
            \end{align}
            where $W_{Q}, W_K, W_V \in \mathbb{R}^{d \times d_k}$ are model parameters. $\bm x_i$ is the input  of this Transformer block. For the first block, it is the vector sum of the table-lookup embedding and the position embedding, i.e.,  $\bm n_{i}+\bm p_{1,i}$; for other blocks, it is the vector sum of the lower Transformer block's output and the position embedding that corresponds to this block.
            
            \medskip
            \underline{\textit{Gating Mechanism.}} After the features are computed by self-attention, we further incorporate with the information of character embeddings. This is given by a gating mechanism based on softmax. For the $i$th word, we compute a control vector $\bm q_i$ from $\bm y_i^{\text{(self)}}$ by a linear transformation. The softmax weight $\bm k_i^\text{(c)}$ for character embedding is given by a linear transformation from $\bm n_i^\text{(c)}$ in Equation~\ref{equation:conv_preprocessing}. The softmax weight $\bm k_i^\text{(y)}$ for Transformer's output is given by another linear transformation from $\bm y_i^{\text{(self)}}$. Then, the gate is computed by
            \begin{align}
                [\alpha_{i, t}^\text{(y)}, \alpha_{i, t}^\text{(c)}]= \operatorname{softmax}\{ \bm q_i^\top \bm k_i^\text{(y)}, \bm q_i^\top\bm k_i^\text{(c)}\}
                \label{gating_softmax}
            \end{align}
            They are used to weigh the feature of the Transformer's layer $\bm v_i^\text{(y)}$ and the feature of character embeddings $\bm v_i^\text{(c)}$, linear transformed from $\bm y_i^\text{(self)}$ and $\bm n_i^\text{(c)}$, respectively.

            \begin{equation}
                \bm h_{i,t}= [\alpha_{i, t}^\text{(y)}\bm v_i^\text{(y)} + \alpha_{i, t}^\text{(c)} \bm v_i^\text{(c)}]
                \label{gating_head}
            \end{equation}
            
                        Similar to Equation~\ref{equation:head}, the output of our gating mechanism is 
            $Y^{\text{(gate)}} = (\bm h_{i, t})_{i,t}$, where $(\cdot)_{i,t}$ represents a block matrix with the elements being $h_{i,t}$.
            
            \medskip
            \underline{\textit{Word Convolution.}}
            Finally, two convolutional layers are applied to the output of the gating mechanism $\bm y^\text{(gate)}_1,\cdots,\bm y^\text{(gate)}_L$ and to extract the local features around each token $\bm y^{\text{(}\text{conv}, l\text{)}}_1,\cdots,\bm y^{\text{(}\text{conv}, l\text{)}}_L$, where $l$ denotes the layer of convolutional layers. The $\bm y^{\text{(conv}, l\text{)}}_i$ is computed by
            \begin{equation}
                \bm y^{\text{(}\text{conv}, l\text{)}}_i = W^{\text{(}\text{conv}, l\text{)}}[\bm y^{\text{(}\text{conv}, l - 1\text{)}}_{i - w};\cdots;\bm y^{\text{(}\text{conv}, l - 1\text{)}}_{i + w}]
            \end{equation}
            where $W^{\text{(}\text{conv}, l\text{)}}$ are the convolution weights, $w = (k - 1) / 2$, and $k$ denotes the window size. In particular, $\bm y^{\text{(}\text{conv}, 0\text{)}}_i$ denotes the output of gating mechanism $\bm y^{\text{(}\text{gate)}}_i$. In these layers, separable convolution~\cite{chollet2017xception} is used. The reason is separable convolution has fewer parameters that is easy for training. For the first and the last words, we add zero padding. Between these layers, we use the $\mathrm{GELU}$ activation function~\cite{hendrycks2016bridging}.
            
            \medskip
            In summary, the NL reader has a few Transformer blocks of self-attention, the gating mechanism, and word convolution. The natural language description is encoded as features $\bm y_1^\text{(NL)},\bm y_2^\text{(NL)},\cdots,\bm y_L^\text{(NL)}$. 
    
    \subsection{AST Reader}
        We design an AST reader to model the structure of the partial AST that has generated. Although our programs are generated by predicting the sequence of grammar rules, these rules alone lack a concrete picture of the program and are insufficient for predicting the next rule. Therefore, our AST reader considers heterogeneous information, including the predicted rules and the tree structures. 
        
         To incorporate such program-specific information, we first represent the code as a sequence of rules, then encode the rules with attention mechanism, and finally use a tree convolution layer to combine the encoded representation of each node with its ancestors.

        \subsubsection{AST Representation}\ 
        
        \medskip
            \underline{\textit{Rule Sequence Embedding.}} To encode the rule information, we use the ID of the rules. Suppose we have a sequence of rules $r_1,r_2,\cdots,r_P$ that are been used to generate the partial AST in a decoding step, where $P$ denotes the length of the sequence. We represent these rules as real-valued vectors $\bm r_1, \bm r_2, \cdots, \bm r_P$ by table-lookup embeddings. 
        
            \medskip
            \underline{\textit{Rule Definition Encoding.}}
            The above table-lookup embedding treats a grammar rule as an atomic token, and loses the information of the rule's content.
            
            To alleviate this problem, we enhance the representation of a rule with the encoding of rule definition.
            
            For a grammar rule $i: \alpha\rightarrow\beta_1\cdots \beta_K$, where $\alpha$ is the parent node and $\beta_1\cdots \beta_K$ are child nodes. They can be either terminal or non-terminal symbols. The index $i$ is the ID of the rule.

            Similar to Equation~\ref{equation:conv_preprocessing}, we encode the rule content as a vector $\bm r^{\text{(c)}}$ by a fully-connected layer with input being the table-lookup embeddings $\bm\alpha, \bm\beta_1, \cdots, \bm\beta_K$ of respective symbols. It is noted that the sequence is also padded to a maximum length. 
            
             Then the rule definition features $\bm{y}_{1}^\text{(rule)},\cdots,\bm{y}^\text{(rule)}_{P}$ are computed by another fully-connected layer as 
            \begin{equation}
                \bm{y}_{i}^\text{(rule)} = W^{\text{(rule)}}[\bm{r}_{i};\bm{r}^\text{(c)};\bm\alpha]
            \end{equation}
            where $\bm r_i$ is the table-lookup embedding of the rule $r_i$, $\bm r_i^\text{(c)}$ is the content-encoding rule representation, and we emphasize the parent node $\alpha$ again. After that, a layer normalization is followed.
            
            \medskip
            \underline{\textit{Position and Depth Embeddings.}}
            Since our AST reader would use self-attention mechanisms, we need to represent the position where a grammar rule is used. 
            
            We first adopt the position embedding as in Equation~\ref{equation:position_embedding}, representing when a rule is used in the sequence $r_1, \cdots, r_P$. The position embeddings are denoted by $\bm p^\text{(r)}_1\cdots,\bm p^\text{(r)}_P$
            
            However, such position embedding does not capture the position of a rule in the AST. 
            We further encode such information by a \textit{depth embedding}.
            If we expand a symbol $\alpha$ by the rule $r: \alpha\rightarrow \beta_1\cdots \beta_K$, we represent the depth of the rule by its parent node, i.e., $\alpha$. 
            In this way, we associate another sequence of table-lookup depth embeddings $\bm{d}_1,\cdots,\bm{d}_P$ to the sequence of used grammar rules $\bm r_1,\cdots,\bm r_P$.

        \subsubsection{Neural Structure of AST Reader.}
            The AST reader is also composed of a stack of blocks ($N_1$ blocks in total). Each block is decomposed into four sub-layers (namely, self-attention, a gating mechanism, NL attention, and tree convolution). We employ a residual connection around each sub-layer except the layer of tree convolution. After each sub-layer, we apply a layer normalization. 
            
            \medskip
            \underline{\textit{Self-Attention.}} To capture the information of AST, we build a Transformer-like self-attention layer, where the input is sum of the rule embedding, position embedding, and depth embedding, i.e., $\bm r_i+\bm d_i+\bm p^\text{(r)}_i$. The self-attention sub-layer extract features $\bm y_1^\text{(ast-self)},\bm y_2^\text{(ast-self)},\cdots,\bm y_P^\text{(ast-self)}$ of AST input, using the same mechanism as Equations~\ref{equation:position_embedding},~\ref{equation:head},~\ref{equation:dot_attention} with different weights but add an additional depth embedding to $\bm p^\text{(r)}_i$. 
            
            \medskip
            \underline{\textit{Gating Mechanism.}} We would like to incorporate the content-encoding rule representation $\bm y_i^\text{(rule)}$ into the Transformer-extracted features. We adopt a gating mechanism as in Equations~\ref{gating_softmax},~\ref{gating_head}, and the fused features becomes $\bm y_1^\text{(ast-g)},\bm y_2^\text{(ast-g)},\cdots,\bm y_P^\text{(ast-g)}$ after this sub-layer.
            
            \medskip
            \underline{\textit{NL Attention.}} During the decoding step, we should be informed of the input NL description. This is given by a multi-head NL attention, similar to the Transformer decoder's attention to its encoder~\cite{vaswani2017attention}. The extracted features are denoted by$\bm y_1^\text{(ast-nl)},\bm y_2^\text{(ast-nl)},\cdots,\bm y_P^\text{(ast-nl)}$.

            \medskip
            \underline{\textit{Tree Convolution.}}
            Should we consider only the above sub-layers, it would be hard for the reader to combine the information of a node with its ancestors. A node can be far away from its ancestors in the rule sequence but is close in structure. Therefore, it is difficult for a traditional Transformer to extract such structural features.
            
            We integrate the features of a node with those of its ancestors. We treat the AST as a graph and use an adjacency matrix $M$ to represent the directed graph. If a node $\alpha_i$ is the parent of $\alpha_j$, then $M_{ji} = 1$. Suppose all the nodes are presented by features $\bm f_1,\cdots, \bm f_n$, their parents' features can be given by the multiplication with the adjacency matrix:
            \begin{equation}
                [\bm{f}^\text{(par)}_{1},\cdots, \bm{f}^\text{(par)}_n] =  [\bm{f}_{1},\cdots,\bm{f}_{n}] M
                \label{equation:tree_conv}
            \end{equation}
            where $\bm{f}^\text{(par)}_{i}$ denotes the parent of the $i$th node. 
            For the father of the root node, we pad it with the feature vector of the root node itself.

            The tree-based convolution window, applied to the current sub-tree, is given by
            \begin{equation}
                \begin{aligned}
                Y^{\text{(tconv, }l\text{)}} &= f(W^{\text{(tconv, }l\text{)}}[Y^{\text{(tconv, }l - 1\text{)}} ;\\& Y^{\text{(tconv, }l - 1\text{)}} M ;\cdots; Y^{\text{(tconv, }l - 1\text{)}} M^{kt-1} ])
                \end{aligned}
            \end{equation}
            where $W^{\text{(tconv, }l\text{)}}$ is the weights of the convolutional layer, $kt$ denotes the window size (we set to 3 in our experiments), $l$ is the layer of these convolutional layers. In particular, $ Y^{\text{(tconv, }0\text{)}} = [\bm y_1^\text{(att)},\bm y_2^\text{(att)},\cdots,\bm y_P^\text{(att)}]$, where $Y^{\text{(tconv, }0\text{)}} \in \mathbb{R}^{d\times P}$. For the last layer of the AST reader, we add additional two convoluation layers. In the equation, $f$ is the activation function and $\mathrm{GELU}$ is applied between these layers.
            
            In summary, the AST reader has $N_1$ blocks of these four sub-layers, and yields the features $\bm y_1^\text{(ast)}, \bm y_2^\text{(ast)},\cdots,\bm y_P^\text{(ast)}$.

    \subsection{Decoder}
        Our final component is a decoder that integrates the information of the generated code with the NL description, and predicts the next grammar rule. Similar to the AST reader, a stack of blocks ($N_2$ blocks in total) each with several sub-layers is used in the decoder as follows. A residual connection is also employed around each sub-layer followed by a layer normalization.  
        
        The decoder takes the non-terminal node to be expanded as a query.
        Inspired by a previous approach~\cite{sun2019grammar}, the querying node is represented as a path from the root to the node to be expanded. For example, if we are going to expand node ``Assign" in Figure~\ref{AST_Example}, the path should be \textit{root, Module, body, Assign}. We represent the nodes in this path as real-valued vectors. Then we apply a fully-connected layer like Equation~\ref{equation:conv_preprocessing} to these vectors and the output of the path (querying node) is $\bm q^\text{(path)}_i$. 
        
        We then apply two attention layers to integrate the outputs of the AST reader and the NL reader. 
        
        We first apply an AST attention layer over the output of the AST reader with queries and extract features $\bm f^\text{(tree)}_1, \cdots, \bm f^\text{(tree)}_P$. 
        In this layer, $Q$ is computed from queries $\bm q^\text{(path)}_1, \cdots, \bm q^\text{(path)}_P$; $K$ and $V$ are computed from the code features $\bm y_1^\text{(ast)},\cdots,\bm y_P^\text{(ast)}$. 
         We further integrate the features from the input description. This integration is also implemented with an NL attention, where $Q$ is computed by feature $\bm f^\text{(tree)}_1, \cdots, \bm f^\text{(tree)}_P$; and $K$ and $V$ are computed by the input description $\bm y_1^\text{(NL)},\cdots,\bm y_L^\text{(NL)}$. 
         
         Finally, a set of two fully-connected layers, where the first layer has a $GELU$ activation function, are followed to extract features for prediction.

    \subsection{Training and Inference}
        We predict the next grammar rule, among all possible candidates, by softmax based on the decoder's last layer features.
        
        We also introduce the pointer network~\cite{see2017get} (essentially, an attention) that can directly copy a token $a$ from the NL description. In this case, the resulting grammar rule is $\alpha\rightarrow a$, where $\alpha$ is a non-terminal symbol to be expanded and $a$ is a terminal symbol. Such pointer mechanism is helpful for user-defined identifiers (e.g., variable and function names).
    
        The choice between softmax rule prediction and the pointer network is given by another gating mechanism $p_g$, also computed from the decoder's last feature. 
        The overall predicted probability of the next grammar rule is
        \begin{equation}
             p(r_i|\cdot) = 
             \begin{cases}
             p_g\ p(r_i|\cdot) & \text{if } i \in \mathbf{D}\\
             (1 - p_\text{g})\operatorname{Pr}\{\text{copy word $t$ at step $i$}|\cdot\}& \text{if } i \in \mathbf{C}
             \end{cases}
        \end{equation}
        where $i$ denotes the ID of the rule, $\mathbf{D}$ is the set of predefined rules, and $\mathbf{C}$ denotes the set of rules in the form of $\alpha\rightarrow a$, where $a$ is a terminal token that occurs in the NL description. 
        
        $p_g$ is the probability of using the type of predefined rules, and the $p(r_i|\cdot)$ (the probability of each predefined rules) are computed by two single-layer perceptrons with the sigmoid and softmax activation functions, respectively, and the input of these layers are the features $\bm h^\text{(dec)}$. 
        
        The pointer network is computed by
        \begin{equation}
            \begin{aligned}
                \xi_t &= \bm v^T\ \mathrm{tanh}(W_{1} \bm h^\text{(dec)} + W_{2} \bm y_{t}^\text{(NL)})\\
                \operatorname{Pr}&\{\text{copy word $t$ at step $i$}|\cdot\} = \frac{\exp{\{\xi_t\}}}{\sum_{j=1}^{L} \exp{\{\xi_j\}}}
            \end{aligned}
            \label{eq:copy}
        \end{equation}
        where $\bm h^\text{(dec)}$ denotes the decoder's last feature. 
        The model is optimized by maximizing negative log likelihood loss against the reference program.
        
        The inference starts with a \textit{start} rule, $start: \textit{snode} \longrightarrow \textit{root}$, expanding a special symbol \textit{snode} to the \textit{root} symbol. The recursive prediction terminates if every leaf node in the predicted AST is a terminal. During prediction, we use beam search with a size of 5. Invalid rules are excluded during beam search.
\section{Evaluation} 
    We evaluated our approach on two types of benchmarks: (1) a Python code generation benchmark, HearthStone, and (2) two semantic parsing benchmarks, ATIS and GEO. 
    
    \subsection{Experiment: HearthStone}
        \paragraph{Dataset.} 
        \begin{figure}
            \centering
              \includegraphics[width=0.9\linewidth]{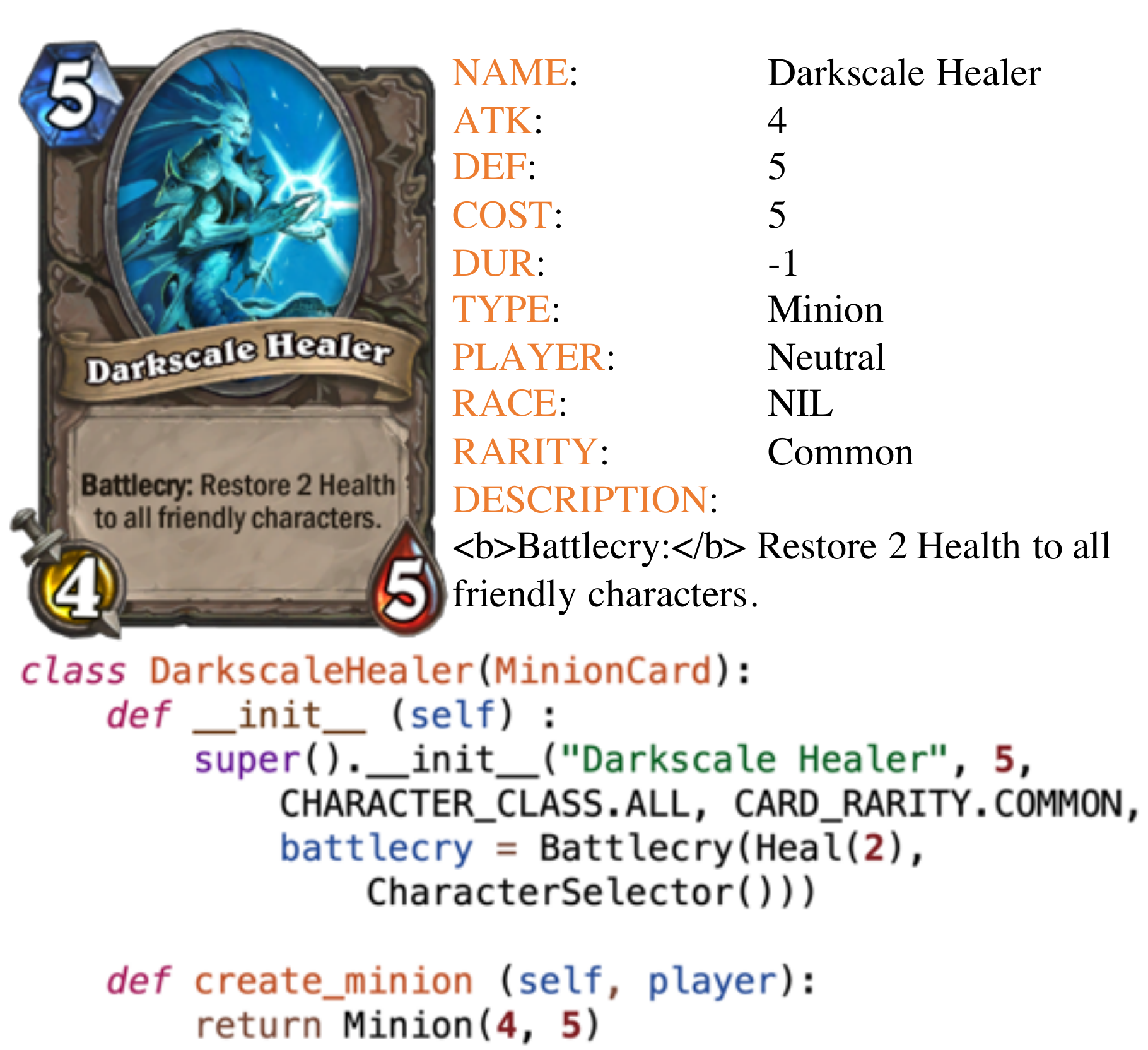}
            \caption{A example of the implement of HearthStone.}
            \label{hs_example}
        \end{figure}
        We first evaluated our approach on the HearthStone benchmark~\cite{ling2016latent}. The benchmark contains Python code that implements 665 different cards of HearthStone. Each card is composed of a semi-structural description and a groundtruth Python program. The Python programs have a length of 84 tokens on average. 
        The description comes with several attributes such as card name, card type, as well as a natural language description for the functionality of the card. A Python program is mainly decided by the natural language description where the attributes decide the constants or identifier names.
        A sample description and its corresponding Python program are shown in Figure~\ref{hs_example}. 
        When preprocessing the card description into token sequences, existing approaches consider two methods. The first~\cite{yin2017syntactic,hayati-etal-2018-retrieval} (called \emph{plain} preprocessing) treats the whole description as plain text and delimit the tokens by standard separators such as space or periods. The second~\cite{rabinovich2017abstract} (called \emph{structural} preprocessing) treats the descriptions as semi-structural and always treat an attribute as one token. In this experiment, we consider both methods and denote the results corresponding to the plain preprocessing as {TreeGen-A} and that corresponding to the structural preprocessing as {TreeGen-B}.
        We followed the train-dev-test split in~\citeauthor{ling2016latent}~(\citeyear{ling2016latent}), and the statistic is listed in Table~\ref{table:stat}.
        
        \paragraph{Metrics.}
        We measured the performance following the metrics in~\citeauthor{sun2019grammar}~(\citeyear{sun2019grammar}). We computed the StrAcc, which is the percentage of programs that has exactly the same token sequence as the ground truth; the BLEU score, which is used to measure the similarity between the generated code and the reference code at the token level; and the Acc+, which is evaluated manually, allows variable renaming on top of StrAcc, for every test case. 
        \paragraph{Settings.}
        For neural networks, we set the number of NL reader layers $N_d = 6$, and $N_1 = N_2 = 5$ for the AST reader as well as the decoder. The size of all embedding is 256. The hidden sizes were all set to the 256 except each fully-connected layers, except the first layer was 1024 dimensions. We applied dropout after each layer (including attention layers, gating mechanism layers, convolutional layers, and fully-connected layers, where the drop rate is 0.15). The model is optimized by Adafactor~\cite{shazeer2018adafactor} with default parameters.
        \begin{table*}[!t]
            \centering
            \resizebox{0.55\textwidth}{!}{
            \begin{tabular}{l|lrrr}
                \toprule
                &\textbf{Model}  &\!\!\!\!\!\!\!\! \textbf{StrAcc} & {\textbf{Acc+}} &\!\!\! {\textbf{BLEU}} \\
                \toprule
                \multirow{2}{*}{\rotatebox{90}{Plain}}
                &LPN~\cite{ling2016latent}  & 6.1 &--& 67.1 \\
                &SEQ2TREE~\cite{dong2016language} &1.5&-- &53.4\\
                &YN17~\cite{yin2017syntactic}& 16.2 &\!\!\!{\tiny$\sim$}18.2& 75.8 \\
                &ASN~\cite{rabinovich2017abstract}& 18.2& -- & 77.6 \\
                &ReCode~\cite{hayati-etal-2018-retrieval} & 19.6 & -- & 78.4 \\
                \cmidrule(lrrr){2-5}
                    &\textbf{TreeGen-A} &\textbf{25.8} &\textbf{25.8} &\textbf{79.3}\\
                \midrule
                \multirow{1}{*}{\rotatebox{90}{Structural}}
                &ASN+SUPATT~\cite{rabinovich2017abstract}& 22.7 & -- & 79.2\\
                &SZM19~\cite{sun2019grammar} & 27.3 & 30.3 & 79.6 \\
                \cmidrule(rrrr){2-5}
                &\textbf{TreeGen-B} &\textbf{31.8} & \textbf{33.3} & 80.8\\
                \cmidrule(rrrr){2-5}
                &\textbf{Location of Structural Convolutional Sub-layer} & & & \\
    &$N_1 = 10, N_2=0$ & 25.8&27.3&80.4\\
                &$N_1 = 10 (7), N_2=0$ & 27.3&28.8&78.5\\
                &$N_1 = 10 (8), N_2=0$ & 25.8&28.8&78.5\\
                &$N_1 = 0, N_2=10$ & 21.2&22.7&79.6\\
                \cmidrule(rrrr){2-5}
                &\textbf{Ablation test} & & &\\
                &Baseline: Transformer & 10.6 ($p=0.015$)&12.1&68.0\\
                &- Tree Convolution &27.3 ($p=0.015$)&27.3 & 80.9\\
                &- Rule Definition Encoding &27.3 ($p<0.001$) & 28.8&\textbf{81.8}\\
                &- Char Embedding & 15.2 ($p<0.001$)& 18.2 & 72.9\\
                &- Self-Attention & 28.8 ($p<0.001$) & 28.8 & 81.0 \\
                \bottomrule        
                
            \end{tabular}}
            \caption{Performance of our model in comparison with previous state-of-the-art results.}
            \label{tab:HS}
        \end{table*}
    
        \paragraph{Overall Results.}
        We show the results in Table~\ref{tab:HS}. In this table, the structural preprocessing has a better performance compared with the plain preprocessing. 
        
         As shown, our model achieves 6 percentage points accuracy improvement with plain preprocessing and 4.5 percentage points accuracy improvement with structural preprocessing. For the BLEU score, our model also achieves the best results. These boosts in performance indicate that TreeGen successfully alleviates the long dependency problem and effectively encodes the structural information in the code generation.
        \paragraph{Time Efficiency.}
            We further evaluated the complexity of our model on the HearthStone, and the result shows that our model is faster than the previous ones. It takes 18s for an epoch on a single Nvidia Titan XP, whereas 180s for the CNN~\cite{sun2019grammar} and 49s for the RNN~\cite{yin2017syntactic}.
        \paragraph{Location of Structural Convolution Sub-layer.}
            One of the keys of our approach is to add the structural convolution sub-layers only to part of the Transformer blocks in the decoder. To evaluate whether this design decision is effective, we evaluate four competing settings: 1) adding the structural convolution sub-layers to all Transformer blocks (i.e., $N_1=10$); 2) adding the structural convolution sub-layers to the first 7 blocks in AST reader (i.e., $N_1=10 (7)$); 3) adding the structural convolution sub-layers to the first 8 blocks in AST reader (i.e., $N_1=10 (8)$); 4) the other adds to none (i.e., $N_1=0$).  
            As we can see, from Table~\ref{tab:HS} our approach adding the sub-layer to all transformer blocks ($N_1=10$) significantly outperforms the last setting ($N_1 = 0$), but slightly worse than the other two settings.
        \paragraph{Ablation Test.} We ablated our model (TreeGen-B was used) to analyze the contribution of each component, results also shown in Table~\ref{tab:HS}. First, we compared our model with the traditional Transformer, which is a Transformer without effective structure modeling. We achieved 21 percentage points higher accuracy (p-value is less than 0.001) and 12 higher BLEU score. 
        This result provides strong evidence of the effectiveness of the AST reader in our model and the importance of the structural information. 
        Next, we replaced the tree convolutional layers in the AST Reader with two layers of fully-connected layers, and we removed the char embedding, rule definition encoding, self-attention layers in turn. The experimental results show the identifiers-encoding, alleviating long-dependency  and structural information significantly influence the accuracy. Please note that in some cases BLEU increases while StrAcc and Acc+ decrease. Here we consider StrAcc and Acc+ more important as they guarantee the correctness of the generated programs and correctness is usually crucial in code generation.
    \subsection{Experiment \uppercase\expandafter{\romannumeral2}: Semantic Parsing}
        \begin{table}[!t]
            \centering
            \small
            \begin{tabular}{llll}
                \toprule
                &&\multicolumn{2}{c}{Exp \uppercase\expandafter{\romannumeral2}}\\
                \cmidrule{3-4}
                \textbf{Statistics} & \textbf{HS}  & {\textbf{ATIS}}  & {\textbf{GEO}}\\
                \midrule
                \# Train & 533  & 4,434  &  600\\
                \# Dev & 66 & 491 &   -\\
                \# Test & 66 & 448  & 280\\
                \midrule
                Avg. tokens in description & 35.0 & 10.6 & 7.4\\
                Max. tokens in description & 76.0 & 48 & 23 \\
                Avg. tokens in code & 83.2 & 33.9 & 28.3\\
                Max. tokens in code & 403 & 113  & 144\\
                \bottomrule
            \end{tabular}
            \caption{Statistics of the datasets we used.}
            \label{table:stat}
        \end{table}

        \paragraph{Dataset.} We further evaluated our approach on the semantic parsing tasks. 
        Our experiment was conducted on two semantic parsing datasets, ATIS and GEO. The input of these datasets is a natural language description, while the output is a short piece of code in lambda calculus. We followed the standard train-dev-test split of these datasets, and the statistics are listed in Table~\ref{table:stat}. 
        
        \paragraph{Metrics and Settings.} In this task, we follow the evaluation of the previous approaches~\cite{dong2016language} and use accuracy as the metric, where the tree exact match was considered to avoid spurious errors. In other words, the order of the children can be changed within conjunction nodes. We followed all the settings in the HS experiment except that we changed the embedding size and the hidden sizes to 128 compared with the setting of the HS experiment.

        \paragraph{Results.}
        \begin{table}[!t]
            \centering
            \resizebox{.9\linewidth}{!}{
            \begin{tabular}{l|lrr}
                \toprule
                & \textbf{Method} & \textbf{ATIS} & \textbf{GEO}\\
                                        \midrule
        \multirow{3}{*}{\rotatebox{90}{Traditional}}            
     
                &ZC07~\cite{zettlemoyer2007online} & 84.6 & 86.1\\
                &FUBL~\cite{kwiatkowski2011lexical} & 82.8 & 88.6\\
                &KCAZ13~\cite{kwiatkowski2013scaling}& - & 89.0 \\ 
                &WKZ14~\cite{wang2014morpho} & \textbf{91.3} & \textbf{90.4}\\
                
                \midrule
        \multirow{3}{*}{\rotatebox{90}{Neural Networks}}        
                &SEQ2SEQ~\cite{dong2016language} & 84.2 & 84.6\\
                &SEQ2TREE~\cite{dong2016language} & 84.6 & 87.1\\
                &ASN~\cite{rabinovich2017abstract} & 85.3 & 85.7\\
                &ASN+SUPATT~\cite{rabinovich2017abstract} & 85.9 & 87.1\\
                &COARSE2FINE~\cite{dong-lapata-2018-coarse} & 87.7 & 88.2\\
                &TRANX~\cite{yin-neubig-2018-tranx} & 86.2 & 88.2\\
                &Seq2Act~\cite{chen-etal-2018-sequence} & 85.5 & 88.9\\
                &Graph2Seq~\cite{xu-etal-2018-exploiting} & 85.9 & 88.1\\
                &SZM19~\cite{sun2019grammar} & 85.0 & - \\
                \cmidrule(rrr){2-4}
                &\textbf{TreeGen} & \textbf{89.1} & \textbf{89.6}\\
                \bottomrule
            \end{tabular}}
            \caption{Accuracy in semantic parsing (in percentage).}
            \label{table:sp}
        \end{table}
        Table~\ref{table:sp} shows the performance of our TreeGen. As seen, the accuracy of our approach is sightly worse than the traditional approach WKZ14~\cite{wang2014morpho}, which is based on the CCG parser and uses a large number of templates. This traditional approach is hard to generalize new datasets. However, our model was directly adopted from the HS dataset, and achieved the highest accuracy, among all neural models~\cite{dong2016language,rabinovich2017abstract,dong-lapata-2018-coarse,chen-etal-2018-sequence,xu-etal-2018-exploiting,sun2019grammar}. This experiment shows the effectiveness and generalizability of TreeGen.

\section{Related Work}
    Code generation achieves significant progress in recent years. The early     
approaches are mainly based on templates \cite{zettlemoyer2007online,zettlemoyer2012learning,kushman2013using,wang2014morpho}. With the prosperity of deep learning, the sequence-to-sequence framework has shown to be effective in various tasks~\cite{sutskever2014sequence}. \citeauthor{ling2016latent}~(\citeyear{ling2016latent}) applied this framework to generate code based on tokens. Unlike natural languages, it is shown that the code contains much more structural information. Thus, the abstract syntax tree (AST) was used in more recent works~\cite{dong2016language,yin2017syntactic,rabinovich2017abstract,hayati-etal-2018-retrieval,yin-neubig-2018-tranx}. However, these studies mainly use recurrent neural networks (RNNs) from the long dependency problem~\cite{bengio1994learning}. \citeauthor{sun2019grammar}~(\citeyear{sun2019grammar}) proposed to use the convolutional neural network (CNN) to handle the long dependency problem. Our approach addresses this problem by Transformer's intensive attention mechanism~\cite{vaswani2017attention}.
    To incorporate the structural information and the idea of self-attention, we propose a tree-based Transformer architecture for code generation.
    
\section{Conclusion}
    In this work, we propose TreeGen for program generation. TreeGen uses the attention mechanism of Transformers to alleviate the long-dependency problem and introduces the AST reader to combine the grammar rules and the AST structure.
    
    The evaluation was conducted on a Python dataset, HearthStone, and two semantic parsing datasets, ATIS and GEO. The experimental results show that our model significantly outperforms existing approaches. We also conducted in-depth ablation tests, which suggests that each component in our model plays a significant role.
\section{Acknowledgments}
This work is sponsored by the National Key Research and Development Program of China under Grant No.~2017YFB1001803, and National Natural Science Foundation of China under Grant
Nos.~61672045, 61529201, and 61922003. Lili Mou is an Amii Fellow; he is supported by the CCAI Chair Program; and he also thanks AltaML for support.

\bibliography{biblio}
\bibliographystyle{aaai}

\appendix

\end{document}